\documentclass{article}
\PassOptionsToPackage{numbers, compress}{natbib}
 \usepackage[preprint]{neurips_2019}

\usepackage[utf8]{inputenc} 
\usepackage[T1]{fontenc}    
\usepackage{hyperref}       
\usepackage{url}            
\usepackage{booktabs}       
\usepackage{amsfonts}       
\usepackage{nicefrac}       
\usepackage{microtype}      
\usepackage{graphicx}
\usepackage{subcaption}
\usepackage{amsmath,amssymb}
\usepackage{float}
\DeclareMathOperator*{\argmax}{\arg\!\max}

\title{Sample-Efficient Reinforcement Learning with Maximum Entropy Mellowmax Episodic Control}

\author{%
  Marta Sarrico\\
  Department of Bioengineering\\
  Imperial College London\\
  \texttt{mvs918@ic.ac.uk} \\
  \And
  Kai Arulkumaran\\
  Department of Bioengineering\\
  Imperial College London\\
  \texttt{kailash.arulkumaran13@imperial.ac.uk} \\
  \And
  Andrea Agostinelli\\
  Department of Bioengineering\\
  Imperial College London\\
  \texttt{aa7918@ic.ac.uk} \\
  \And
  Pierre Richemond\\
  Data Science Institute\\
  Imperial College London\\
  \texttt{phr17@ic.ac.uk} \\
  \And
  Anil A. Bharath\\
  Department of Bioengineering\\
  Imperial College London\\
  \texttt{a.bharath@imperial.ac.uk}
}

\begin{document}

\maketitle

\begin{abstract}
Deep networks have enabled reinforcement learning to scale to more complex and challenging domains, but these methods typically require large quantities of training data. An alternative is to use sample-efficient episodic control methods: neuro-inspired algorithms which use non-/semi-parametric models that predict values based on storing and retrieving previously experienced transitions. One way to further improve the sample efficiency of these approaches is to use more principled exploration strategies. In this work, we therefore propose maximum entropy mellowmax episodic control (MEMEC), which samples actions according to a Boltzmann policy with a state-dependent temperature. We demonstrate that MEMEC outperforms other uncertainty- and softmax-based exploration methods on classic reinforcement learning environments and Atari games, achieving both more rapid learning and higher final rewards.
\end{abstract}

\section{Introduction}
Despite the successes of deep reinforcement learning (DRL) agents \cite{arulkumaran2017deep}, these models have a sample-efficiency limitation: DRL agents typically require hundreds of times more experience than a human to reach similar levels, suggesting a large gap between current DRL algorithms and the operation of the human brain \cite{yu2018towards,lake2017building}. Recently, new neuro-inspired episodic control (EC) algorithms have demonstrated rapid learning, as compared to state-of-the-art DRL methods \cite{blundell2016model,pritzel2017neural}. These algorithms were inspired by human long-term memory, which can be divided into semantic and episodic memory: the former is responsible for storing general knowledge and facts about the world, whilst the latter is related to recollecting our personal experiences \cite{greenberg2010interdependence}. EC, introduced by Lengyel \& Dayan \cite{lengyel2008hippocampal}, is inspired by this biological episodic memory, and models one of the several different control systems used for behavioural decisions as suggested by neuroscience research \cite{daw2005uncertainty}. As opposed to other RL systems, EC enables rapidly learning a policy from sparse amounts of experience.

An alternative factor in the sample-efficiency of RL methods is the existence of an effective exploration policy that is able to collect diverse experiences from the environment to learn from. The deep $Q$-network (DQN) \cite{mnih2015human}, a notable DRL algorithm, as well as current EC algorithms \cite{blundell2016model,pritzel2017neural}, use the naive $\epsilon$-greedy exploration policy \cite{sutton1998introduction}. More principled exploration methods, such as upper confidence bound (UCB) \cite{auer2002finite,chen2017ucb} and Thompson sampling (TS) \cite{osband2013more}, sample actions based on uncertainty over their consequences. Another alternative is to sample from a distribution over actions, such as the Boltzmann distribution \cite{sutton1998introduction}, performing importance weighting of actions proportionally to the Gibbs-Boltzmann weights of their state-action values.
Yet another solution is to use an alternative objective, yielding a policy that balances between maximising both the expected return and its entropy over states \cite{asadi2017alternative,kimdeepmellow}. Similarly to supervised learning settings, the latter approach corresponds to an entropic regularisation over the solution \cite{fox2015taming,nachum2017bridging,neu2017unified,richemond2017short}.

In this work, we hence propose maximum entropy mellowmax episodic control (MEMEC), which combines EC models with the maximum entropy mellowmax policy \cite{asadi2017alternative} for principled exploration. Without resorting to maximum entropy RL, hence decoupling exploration benefits from the objective, we show that this softmax-based exploration strategy can still improve both the sample-efficiency and final returns of EC methods, while retaining their simple, low-bias Monte Carlo returns. We test MEMEC on a wide variety of domains, where it shows improvements over the original EC methods, as well as the same EC methods with alternative exploration policies.
\section{Background}

\textbf{Reinforcement Learning:} The RL setting is formalised by Markov decision processes (MDPs). MDPs are characterised by a tuple $\langle S,A,R,T,\gamma \rangle$, where $S$ is the set of states, $A$ is the set of actions, $R$ is a reward function which is the immediate, intrinsic desirability of a certain state, $T$ is the transition dynamics, and $\gamma \in [0,1)$ is a discount factor that trades off the value of current and future rewards. The goal of RL is to find the optimal policy, $\pi^*$, that maximizes the expected cumulative discounted return when followed from any state $s \in S$.

$Q$-learning \cite{watkins1992q}, a widely used temporal difference (TD) method, can learn value functions by bootstrapping. The DQN uses $Q$-learning, updating a neural-network-based state-action value function, $Q^{\pi}(s,a; \theta)$, parameterised by $\theta$. As a baseline across all environments we used a strong DRL method---the dueling double DQN (D3QN) \cite{mnih2015human,van2016deep,wang2015dueling} (further detailed in Section \ref{sec:ec}).

\begin{figure}[ht]
\centering
\includegraphics[width=.7\columnwidth]{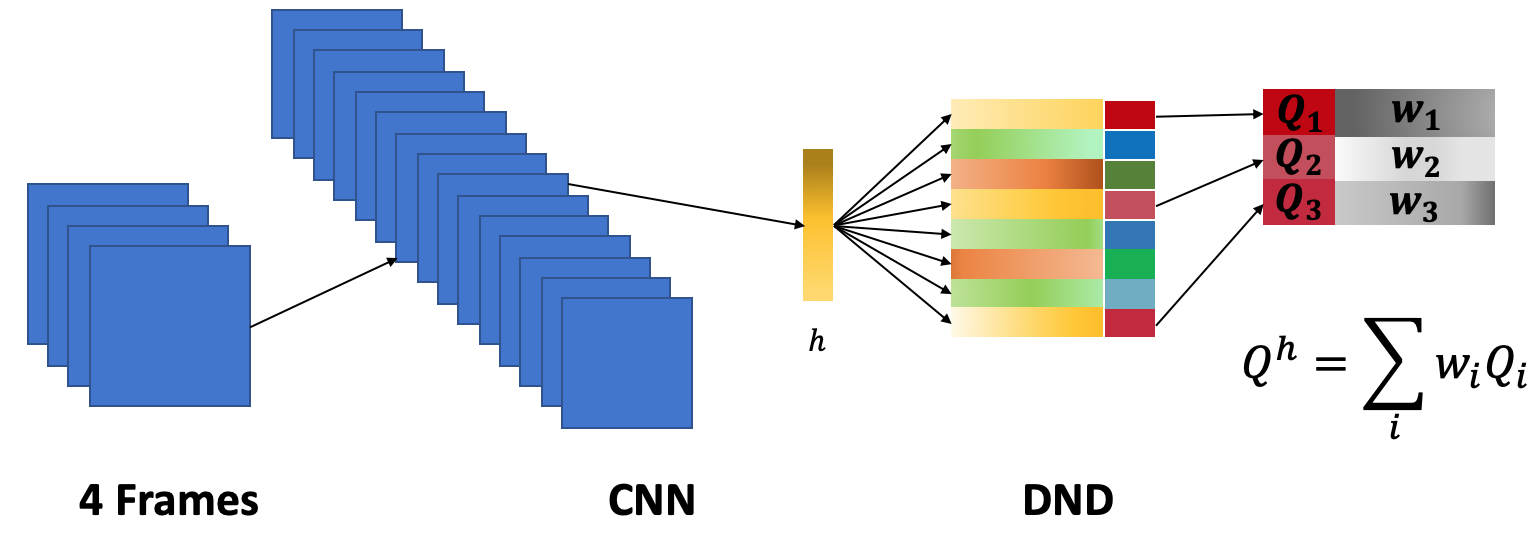} 
\caption{Architecture of NEC (for Atari games): a convolutional neural network receives a state (4 frames of the game) as input and outputs a key $h$, an embedding of that state. Lookup: Key $h$ is compared to the stored keys in the differentiable neural dictionary (DND) and the $k$ most similar ones ($k=3$ in this case) are retrieved. $Q^h$ is computed as a weighted sum of the $Q$-values associated with the retrieved keys.}
\label{NEC_architecture}
\end{figure}

\textbf{Episodic Control:} EC models use a memory buffer that stores state-action pairs and their associated episodic returns, $(s, a, R)$, with control performed by replaying the most rewarding actions based on the similarity between the current and stored states. Implementations of these EC methods include the non-parametric model-free EC (MFEC) \cite{blundell2016model}, and the semi-parametric neural EC (NEC) \cite{pritzel2017neural}. Further details on MFEC and NEC can be found in Figure \ref{NEC_architecture} and Section \ref{sec:ec}.

\section{Exploration Strategies}

For both MFEC and NEC, in order to trade off exploration and exploitation in the environment, the agent follows an $\epsilon$-greedy policy. This consists of a sampling strategy where the agent either uniformly picks over all actions with probability $\epsilon$, or instead chooses the action that maximises the predicted return.

\textbf{Uncertainty-based:} UCB and TS are two exploration strategies that sample actions based on predicted uncertainties over the value of the return, and hence require learning a distribution over each $Q$-value. Common practice is to assume a (sub-)Gaussian distribution over $Q$-values, which is both convenient and yields tight regret lower bounds \cite{lairobbins}.

In UCB, the action that is chosen at each timestep is the one that is the maximum over the sum of the value (estimated mean of the distribution) and its uncertainty $U(s,a)$ (a function of the estimated standard deviation of the distribution): $\argmax_{a \in \mathcal{A}}Q(s,a)+U(s,a)$. TS can be implemented by sampling from a posterior over $Q$-functions: ${\tilde{Q}(s, a) \sim \mathcal{N}(Q(s, a), \tilde{\sigma}(Q(s, a))\quad \forall a \in \mathcal{A}}$, where $\tilde{\sigma}(Q(s, a))$ is the estimated standard deviation of the $Q$-value, and we are assuming independence of state-action value functions. Then, the $\argmax$ action is chosen greedily from the sampled $Q$-function: $\argmax_{a \in \mathcal{A}} \tilde{Q}(s,a)$. The difficulty with these methods is that they require learning a probabilistic $Q$-function, and learning well-calibrated uncertainties using neural networks is still a challenging problem \cite{osband2016risk}. Furthermore, the actions are sampled independently per timestep, and so the resulting policy can be inconsistent over time \cite{osband2016deep}.

\textbf{Softmax-based:} An alternative which does not have these disadvantages is to use a softmax-based policy, which only requires learning the expected $Q$-values. By applying the Boltzmann operator (with inverse temperature $\beta$) to the $Q$-values, $\operatorname{boltz}_{\beta}(Q)=\frac{\sum_{i=1}^{n} Q_{i} e^{\beta Q_{i}}}{\sum_{i=1}^{n} e^{\beta Q_{i}}}$, one can sample actions from the resulting Boltzmann distribution over actions, where more promising actions will be sampled more frequently. However, the Boltzmann operator is \emph{not} a non-expansion (it is \emph{not} Lipschitz with constant $< 1$)  for all values of the temperature parameter \cite{littman1996generalized}; critical to the proof of convergence in TD learning, the non-expansion property is a sufficient condition to guarantee convergence to a unique fixed point. As a solution, Asade \& Littman \cite{asadi2017alternative} introduced the mellowmax operator (with hyperparameter $\omega$),
$\operatorname{mm}_{\omega}(Q)=\frac{\log \left(\frac{1}{n} \sum_{i=1}^{n} e^{\omega Q_{i}}\right)}{\omega}$, as an alternative softmax function for value function optimisation, which is a non-expansion in the infinity norm for all temperatures.

As the mellowmax operator can return the maximum (as $\omega \rightarrow \infty$) or minimum (as $\omega \rightarrow -\infty$) of a set of values, Asadi \& Littman \cite{asadi2017alternative} also proposed optimising $\omega$ under a maximum entropy constraint. This results in the maximum entropy mellowmax policy: $\pi_{mm}(a | s)=\frac{e^{\beta Q(s, a)}}{\sum_{a \in \mathcal{A}} e^{\beta Q(s, a)}}$, which is of the same functional form as the Boltzmann policy, but where the optimal $\beta$ can be solved for using a root-finding algorithm, such as Brent's method \cite{brent2013algorithms}. The equation which is solved for is $\sum_{a \in \mathcal{A}} e^{\beta\left(Q(s, a)-\operatorname{mm}_{\omega}(Q(s, \cdot))\right)}\left(Q(s, a)-\operatorname{mm}_{\omega}(Q(s, \cdot))\right)=0.$

MEMEC, which uses the maximum entropy mellowmax policy for exploration, hence benefits from softmax-based exploration, where actions are chosen as a function of their estimated value, with a state-dependent temperature on the Boltzmann distribution over actions. As we show in our experiments, this approach outperforms both the original EC methods with $\epsilon$-greedy exploration, as well as the alternative exploration methods outlined here. We do not train on a maximum entropy objective, thereby purely using the maximum entropy mellowmax policy for exploration.

\section{Experiments}\label{sec:experiments}

\textbf{Environments:} We evaluated EC methods with different exploration strategies, as well as a D3QN baseline, in three sets of domains. Firstly, CartPole and Acrobot, two classic RL problems, implemented in OpenAI Gym \cite{brockman2016openai}. Secondly, OpenRoom, and FourRoom, gridworld domains re-implemented from Machado et al. \cite{machado2017laplacian}. For these domains we trained the agents for 100,000 steps, and evaluated them every 500 steps. Thirdly, Pong, Space Invaders, Q*bert, Bowling, and Ms. Pac-Man, video games with high-dimensional visual observations from the Atari Learning Environment \cite{bellemare2013arcade}. Due to computational resource limitations, Atari training was performed for 5,000,000 steps (20,000,000 frames) for MFEC and 3,500,000 steps (12,000,000 frames) for NEC; evaluation was performed every 100,000 steps. For Atari games, we use standard preprocessing for the state \cite{mnih2015human}, and similarly repeat actions for 4 game frames (= 1 agent step). We used Gaussian random projections for MFEC, as this obtained better results than the variational autoencoder embeddings in the original work \cite{blundell2016model}. For all experiments we report the mean and standard deviation of each method, calculated over three random seeds.

\textbf{Hyperparameters:} For all EC methods, we primarily used the original hyperparameters \cite{blundell2016model,pritzel2017neural}, but chose to keep some consistent across MFEC and NEC for consistency. These were to use the more robust inverse distance weighted kernel (Equation \ref{kernel}) for MFEC and NEC, $k$ = 11 nearest neighbours, and a discount factor $\gamma = $ 0.99 across all domains. For the episodic memories, the buffer size was set to 150 for the room domains, 10,000 for the classic control domains and 100,000 (due to computational resource limitations) for the Atari games. We used the original hyperparameters for D3QN on Atari games \cite{wang2015dueling}, and tuned them manually for the other domains. The full details are present in Section \ref{sec:hyperparameters}.

\textbf{$\omega$ Hyperparameter:} The $\omega$ hyperparameter in the mellowmax operator requires tuning per domain \cite{asadi2017alternative,kimdeepmellow}. With high $\omega$, ${mm}_\omega$ acts as a max operator and with $\omega$ approaching zero ${mm}_\omega$ acts as a mean operator. Thus, $\omega$ should not be too large, as the agent will act greedily, nor too small, as the agent will behave randomly. To tune this hyperparameter we implemented a grid search with the following $\omega$ values: for the Gridworld and classic Control environments, $\omega \in\{5, 7, 9, 12\}$; for Atari domain, $\omega \in\{10, 20, 25, 30, 40, 50, 60\}$, depending on the game. As in prior work \cite{asadi2017alternative,kimdeepmellow}, we show results for MEMEC with the best $\omega$. We set $\omega$ to 7.5 for all non-Atari domains, 25 for Pong, 40 for Q*Bert and Space Invaders, 50 for Bowling and 60 for Ms. Pac-Man.

\textbf{Uncertainty-based Exploration:} For the five simpler domains, we tested $\epsilon$-greedy, UCB, TS, Boltzmann and maximum entropy mellowmax exploration strategies. While UCB sometimes performed well, TS performed extremely badly. Upon closer examination, the uncertainty estimates over the $Q$-values as calculated via the covariance matrix defined over the inverse distance kernel \cite{pritzel2017neural} embeddings over the nearest neighbour keys were both relatively constant, meaning that UCB was close to $\epsilon$-greedy with small $\epsilon$, and relatively large, causing TS' poor performance (high variance samples for the $Q$-values). We were unable to improve the uncertainty estimates by fitting a Gaussian process to the nearest neighbours and optimising over $\delta$ in the inverse distance weighted kernel. Theoretically, UCB and TS often have similar frequentist regret bounds \cite{lattimore2018bandit}, but our experiments demonstrate how dependent these are on well-calibrated uncertainties.\footnote{Despite this, we have retained the results for UCB in any experiments conducted with this method.} Due to computational limitations, and as a result of these experiments on the simpler domains, we only evaluated $\epsilon$-greedy and mellowmax, as well as the D3QN baseline, on the Atari games.

\section{Results}

\begin{figure}[ht]
\centering
\includegraphics[width=.53\columnwidth]{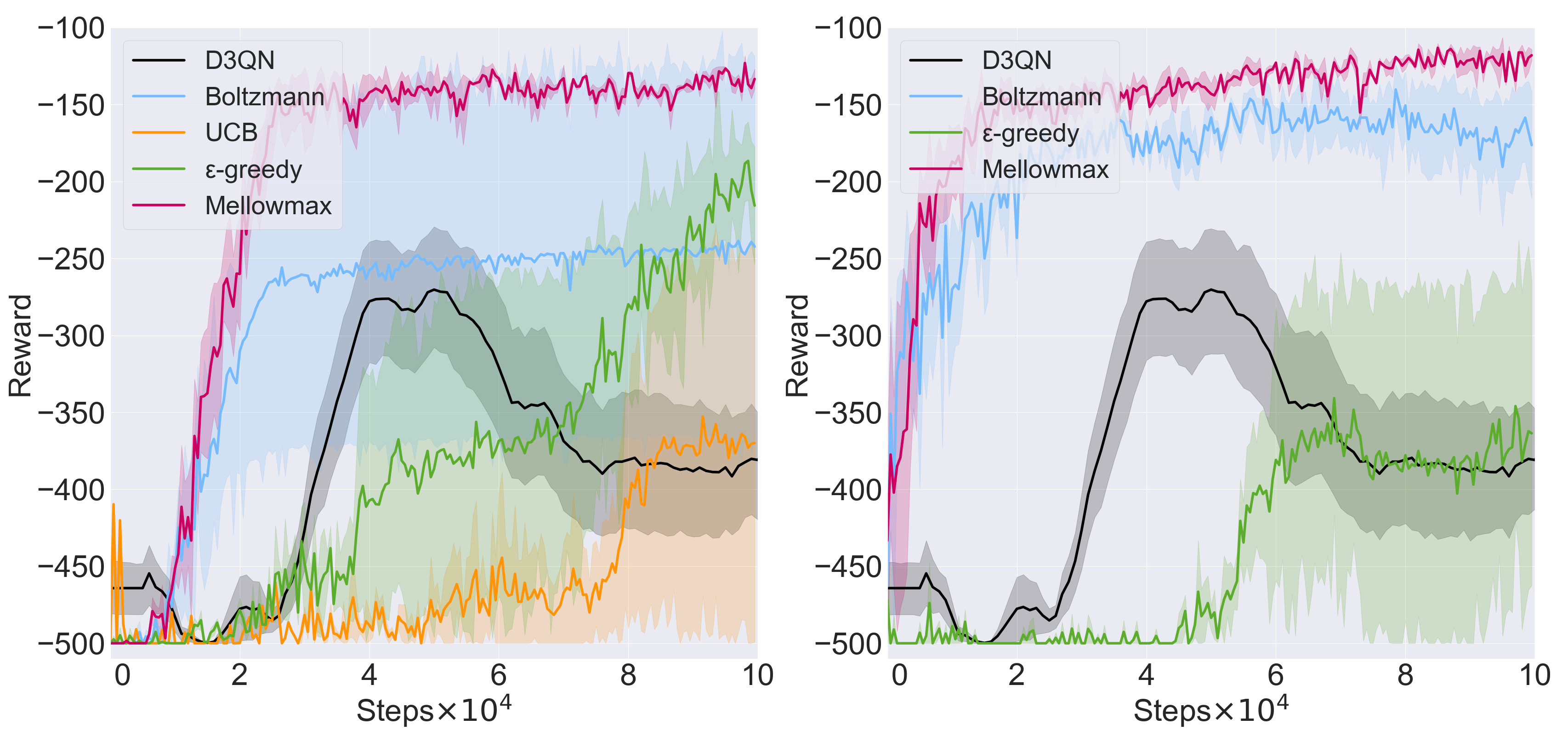} 
\caption{Learning curve on Acrobot with MFEC-based MEMEC.}
\label{acrobot}
\end{figure}

\textbf{Classic Control:} All five methods solved Cartpole (see Section \ref{sec:cartpole results}), as it is a relatively simple environment where even purely random exploration can reach highly rewarding states. In Figure \ref{acrobot} we show the methods' performance in Acrobot, using both MFEC and NEC. In both cases, MEMEC outperformed the other exploration strategies, not only getting to higher final rewards, but starting to learn faster. The next best strategy---the Boltzmann policy---had high variance over random seeds.
As discussed previously in Section \ref{sec:experiments}, UCB performed poorly, which we believe is due to poorly calibrated uncertainties. The D3QN baseline learned more slowly, and converged to a suboptimal policy.

\textbf{Gridworld:} MFEC performed best with softmax-based exploration strategies, whilst nearly all methods tended to perform poorly with NEC (see Section \ref{sec:room results}). Whilst MFEC with $\epsilon$-greedy was able to solve OpenRoom, it had high variance for FourRoom. In contrast to most of our other results, NEC performed best with $\epsilon$-greedy, but still had poor performance on FourRoom. D3QN was not able to consistently solve OpenRoom, and failed to solve FourRoom.

\begin{table*}[t]
\centering
\caption{Final averaged rewards for the Atari games: the values indicate the mean of the last 5 evaluations, averaged over 3 initial random seeds, and their corresponding standard deviations.}
\scriptsize
\begin{tabular}{ c c c | c | c c | c}
\hline
  \textbf{Environments} & \multicolumn{2}{c}{\textbf{MFEC}} & \multicolumn{1}{c}{\textbf{D3QN}}& \multicolumn{2}{c}{\textbf{NEC}}& \multicolumn{1}{c}{\textbf{D3QN}}\\
 & $\epsilon$-greedy  & Mellowmax  & & $\epsilon$-greedy  & Mellowmax \\
 \hline
 \hline 
 
  {Bowling} & $62 \pm{.8}$& $\pmb{68 \pm.7}$&$26 \pm{12}$& $11 \pm{9}$& $9 \pm{7}$ &  $\pmb{22 \pm16}$\\
  
  {Q*Bert} & $3896 \pm{710}$& $\pmb{11610 \pm898}$&  $3743 \pm{1100}$&$3951 \pm{1321}$ & $\pmb{5654 \pm483}$&$1480 \pm{271}$ \\
 
  {Ms. Pac-Man} & $4178 \pm{510}$& $\pmb{6968 \pm787}$& $2101 \pm{56}$& $3900 \pm{852}$& $\pmb{6997 \pm256}$&$1851 \pm{98}$ \\
 
  {Space Invaders} & $672 \pm{13}$ & $\pmb{1029 \pm157}$& $737 \pm{29}$& $598 \pm{110}$ & $\pmb{916 \pm228}$&$756 \pm{30}$\\

 {Pong} & $\pmb{17 \pm2}$& $7 \pm{4}$&  $6 \pm{4}$ & $-7 \pm{.9}$ & $-11 \pm{.5}$& $\pmb{-5 \pm6}$\\
 \hline
\end{tabular}
\label{table2}
\end{table*}

\begin{figure}[ht]
\centering
\includegraphics[width=1\columnwidth]{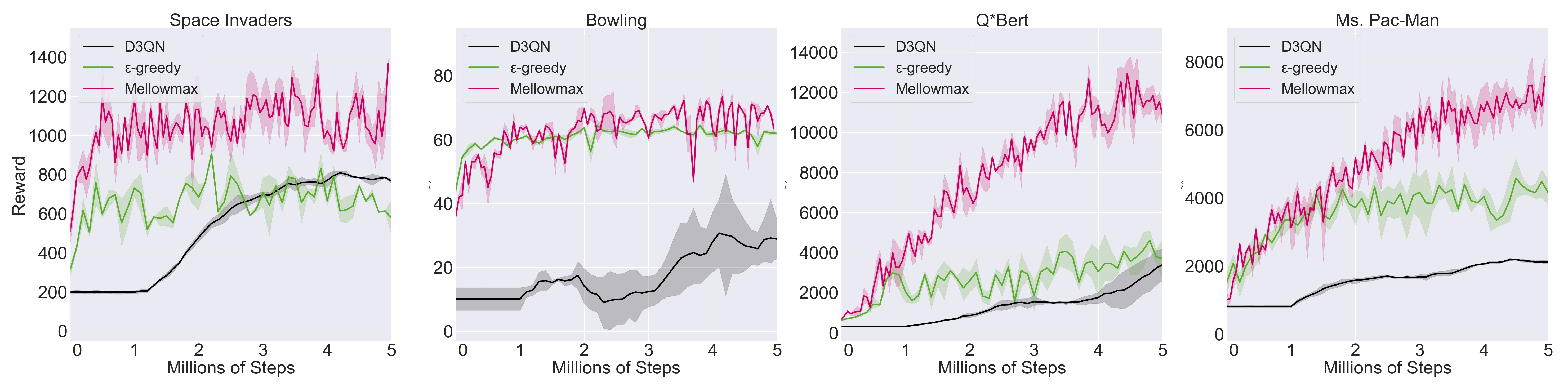} 
\caption{Learning curves for MFEC-based MEMEC agents on 4 Atari games.}
\label{MFEC_Atari}
\end{figure}

\textbf{Atari:} Figure \ref{MFEC_Atari} and Table \ref{table2} show the performance of our method compared to the D3QN baseline and to the $\epsilon$-greedy policy with EC for five different Atari games. Overall, MEMEC outperformed the other methods in these games most of the time, not only in terms of the maximum achieved reward, but also in terms of the learning speed (see Section \ref{sec:Atari results} for additional learning curves). For Space Invaders,  Q*Bert and Ms. Pac-Man, our MEMEC agents, for both EC methods, outperformed the other algorithms, showing more rapid learning and higher final scores. For Bowling, MFEC-based MEMEC performed similarly to $\epsilon$-greedy, and somewhat worse in Pong. NEC-based agents---either with $\epsilon$-greedy or mellowmax policies---failed to solve Bowling and Pong. We note that due to computational limitations we were unable to use the very large buffer sizes from the original works, which can also explain the inability to reproduce the performance of $\epsilon$-greedy EC; it remains to be seen if using larger buffer sizes would similarly improve the performance of MEMEC.

\section{Discussion}

In this work we have investigated the use of more principled exploration methods in combination with EC, and proposed MEMEC, a method that addresses one of the main limitations of state-of-the-art RL models, namely, sample efficiency. We show over a range of domains, including Atari games, that MEMEC can reliably achieve higher rewards faster, with stable performance across seeds.


One limitation of mellowmax-based policies is its sensitivity to the value of $\omega$ across different domains \cite{asadi2017alternative,kimdeepmellow}, and the subsequent searches used to find optimal values, which are prohibitive in domains such as Atari. Implementing methods to automatically determine and tune $\omega$ will be an important area for future work.

\bibliography{bibliography}

\newpage

\section{Additional Background}
\label{sec:ec}

\textbf{Deep Q-network:} The DQN uses $Q$-learning, updating a neural-network based state-action value function, $Q(s,a; \theta)$, parameterised by $\theta$.

The deep Q-network receives the state $s$ as input and outputs a vector with the state-action values $Q(s,a,\theta) \quad \forall a \in \mathcal{A}$. In order to improve the stability of RL with function approximators, the original authors proposed using experience replay \cite{lin1992self} and a target network \cite{mnih2015human}. The target network, parameterised by $\theta^-$, is equal to an older version of the online network and it is updated every $\tau$ steps. The TD target for the DQN is defined by:

\begin{equation}\label{dqn}
y^{DQN}_t = r_{t+1}+\gamma \max _{a}Q(s_{t+1}, a; \theta_{t}^-).
\end{equation}

The experience replay is a cyclic buffer that stores the $(s_t, a_t, r_{t+1}, s_{t+1})$ transitions that the agent observes. Samples are retrieved from it uniformly at regular intervals to train the network.

There have been several proposed extensions that improve this algorithm. Firstly, the $\max$ operator present in $Q$-learning is responsible for both the selection and evaluation of state-action values, and hence can result in overoptimistic estimated $Q$-values. In double $Q$-learning \cite{hasselt2010double}, an alternative update rule for $Q$-learning, the selection of actions and the evaluation of their values are performed by separate estimators. To do so, two value functions are implemented, parameterised by $\theta$ and $\theta'$: One is used for the greedy selection of actions whereas the other is used to determine the corresponding value. For the double DQN algorithm, the online and target networks are used as the two estimators \cite{van2016deep}. In this case, the TD target can be written as:
\begin{equation}\label{ddqn}
y^{DDQN}_t = r_{t+1}+\gamma Q(s_{t+1},\argmax_{a}Q(s_{t+1}, a; \theta_{t});\theta_{t}^-).
\end{equation}
Another extension based on the DQN is the dueling double DQN (D3QN) algorithm \cite{wang2015dueling}.
The dueling architecture extends the standard DQN by calculating separate advantage $A_{\theta}$ and value $V_{\theta}$ streams, sharing the same convolutional neural network as a base. This imposes the form that state-action values can be calculated as offsets (``advantages'') from the state value. The $Q$-function is then computed as:
\begin{equation}\label{d3qn}
Q_{\theta}(s, a) =V_{\theta}(s)+(A_{\theta}(s, a)-\frac{1}{|\mathcal{A}|} \sum_{a^{\prime}} A_{\theta}(s, a^{\prime})).
\end{equation}

\textbf{Model-free episodic control:} In MFEC, the episodic controller $Q^{EC}(s,a)$ is represented by a fixed-size table for each action. Each entry corresponds to the state-action pair and the highest return ever obtained from taking action $a$ from state $s$. Whenever a new state is encountered, the return is estimated based on the average of the observed returns from the $k$-nearest states $s^{(i)}$: 

\begin{equation}\label{mfec}
\widehat{Q^{\mathrm{EC}}}(s, a)=\left\{\begin{array}{ll}{\frac{1}{k} \sum_{i=1}^{k} Q^{\mathrm{EC}}\left(s^{(i)}, a\right)} & {\text { if }(s, a) \notin Q^{\mathrm{EC}}}, \\ {Q^{\mathrm{EC}}(s, a)} & {\text { otherwise }}.\end{array}\right.
\end{equation}

The state-action value function $Q^{EC}(s,a)$ is updated at the end of each episode, replacing the least-recently-used transitions with new state-action pairs and their corresponding episodic returns. If the pair $(s,a)$ already exists in the table, the stored value will be updated to the maximum of the stored return and the newly observed return.

For high-dimensional state spaces, the original authors used Gaussian random projections in order to reduce the dimensionality of the states before using them in the episodic controller.

\textbf{Neural Episodic Control:} NEC consists of two components: a neural network for learning a feature mapping, and a differentiable neural dictionary (DND) per action (see Figure \ref{NEC_architecture}). The neural network receives as input the state $s$ and outputs a key/embedding $h$. Each DND memory performs a lookup based on the current key $h$, by comparing it with the keys $h_i$ already stored in the DND, and retrieving the $k$ most similar keys with their corresponding $Q_i$ values. The predicted $Q(s,a)$ value, based on $h$, is the weighted sum of the retrieved $Q_i$ values: $\sum_{i} w_{i} Q_{i}$. Each weight $w_i$ is calculated by using the inverse distance weighted kernel:
\begin{equation}\label{kernel}
k\left(h, h_{i}\right)=\frac{1}{\left\|h-h_{i}\right\|_{2}^{2}+\delta}   .
\end{equation}

Updating the DND involves appending new key-value pairs to the dictionary or, for cases where the same state already exists in that dictionary, updating the value of $Q(s,a)$ as in $Q$-learning:
\begin{equation}\label{upd_dnd}
Q_{i} \leftarrow Q_{i}+\alpha\left(R^{(n)}(s, a)-Q_{i}\right),
\end{equation}
where $\alpha$ is the learning rate, and $R^{(n)}$ is the $n$-step return \cite{sutton1998introduction} bootstrapped from the predicted $Q$-values from NEC during each episode.

The parameters of the neural network are trained by minimising the mean squared error loss between the predicted $Q$-values and the stored returns, using a random sample of $\left(s_{t}, a_{t}, R^{(n)}_t\right)$ tuples stored in experience replay \cite{lin1992self}, in a similar fashion to the DQN algorithm.

\newpage

\section{Cartpole Results}
\label{sec:cartpole results}

\space
\begin{figure}[ht]
  \begin{subfigure}[b]{0.48\textwidth}
    \includegraphics[width=\textwidth]{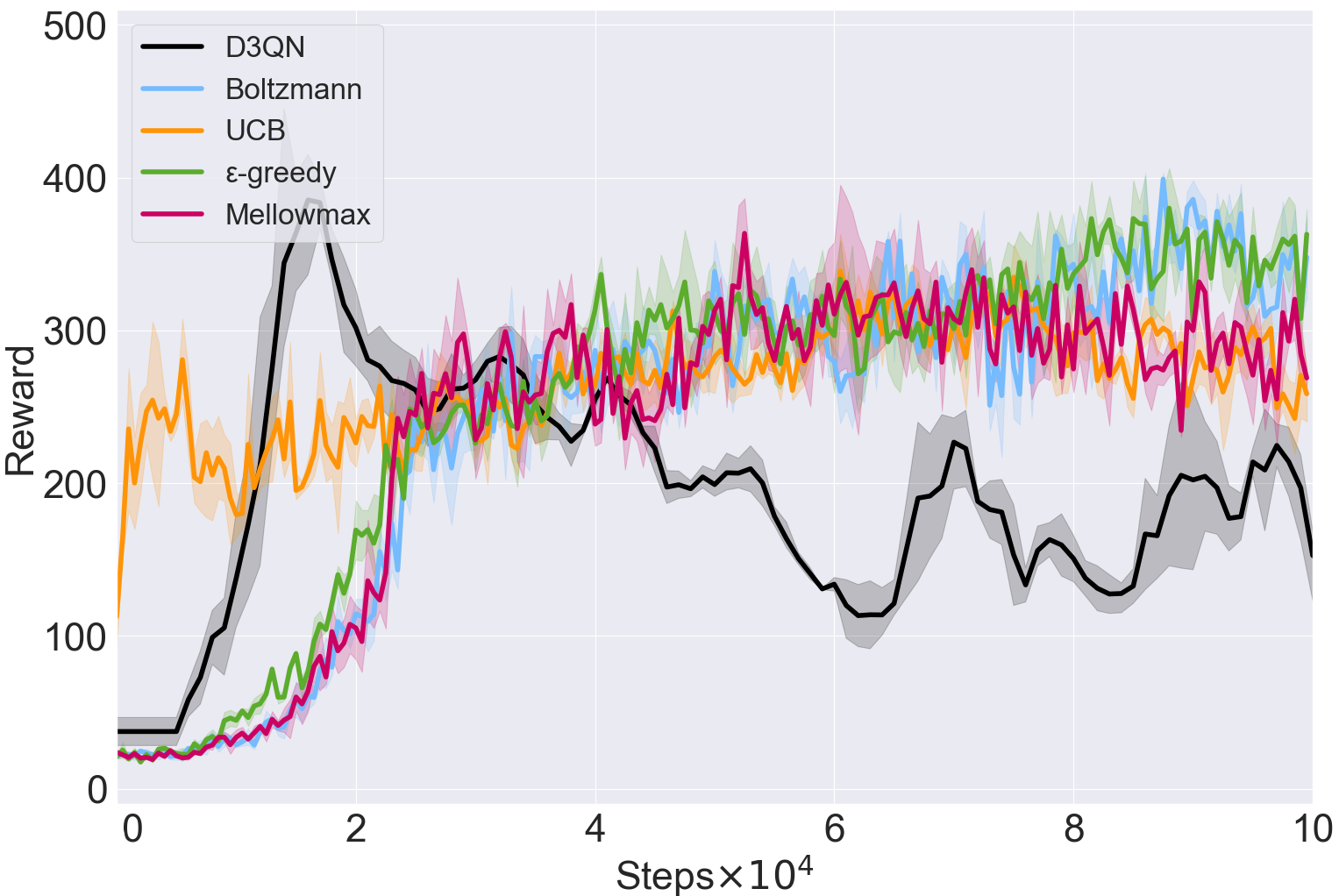}
    \caption{MFEC-based algorithms.}
  \end{subfigure}
  \begin{subfigure}[b]{0.48\textwidth}
    \includegraphics[width=\textwidth]{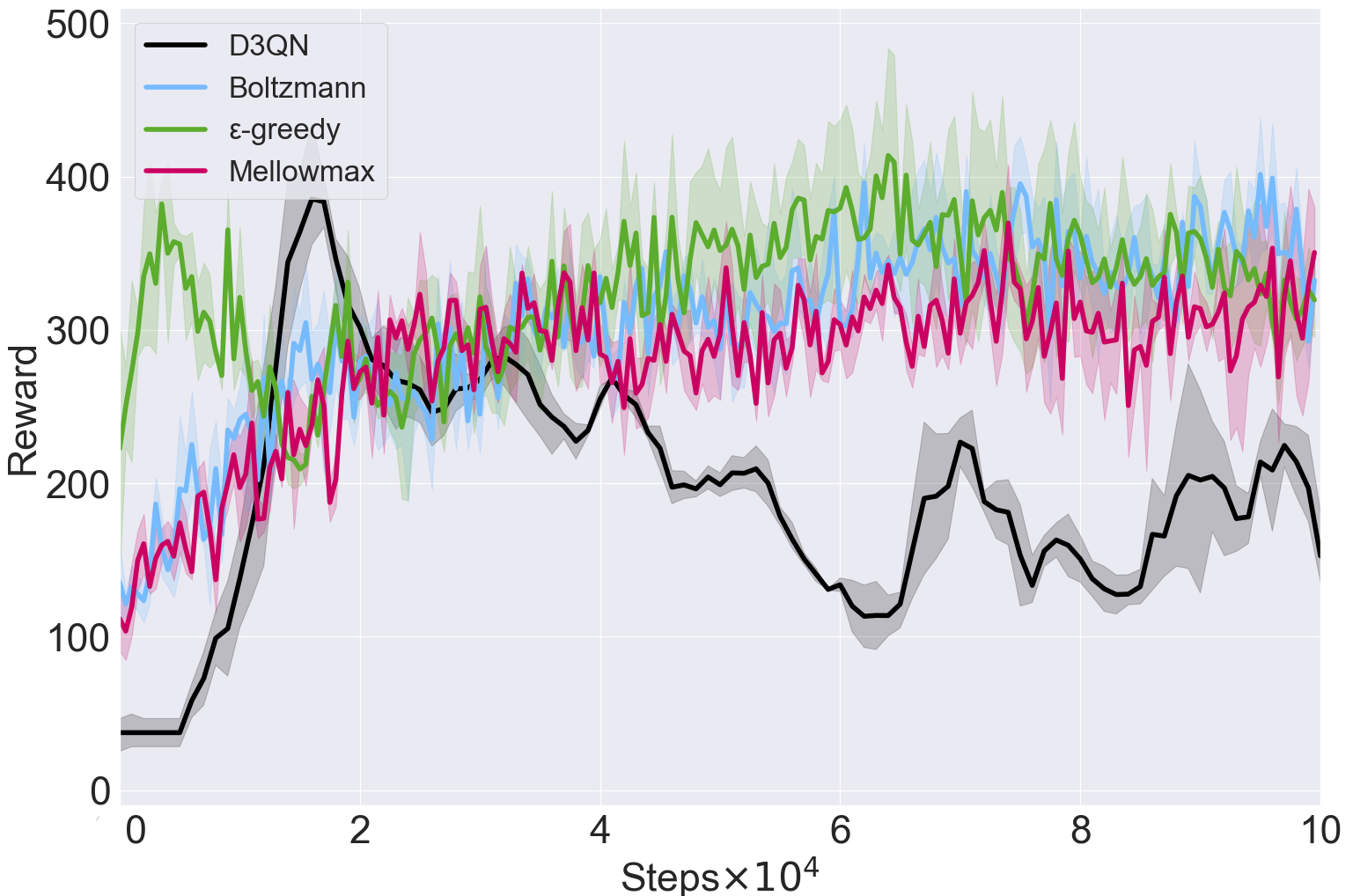}
    \caption{NEC-based algorithms.}
  \end{subfigure}
  \caption{Learning curves on Cartpole.}
\end{figure}

\newpage

\section{GridWorld Results}
\label{sec:room results}

\space
\begin{figure}[ht]
  \begin{subfigure}[b]{0.48\textwidth}
    \includegraphics[width=\textwidth]{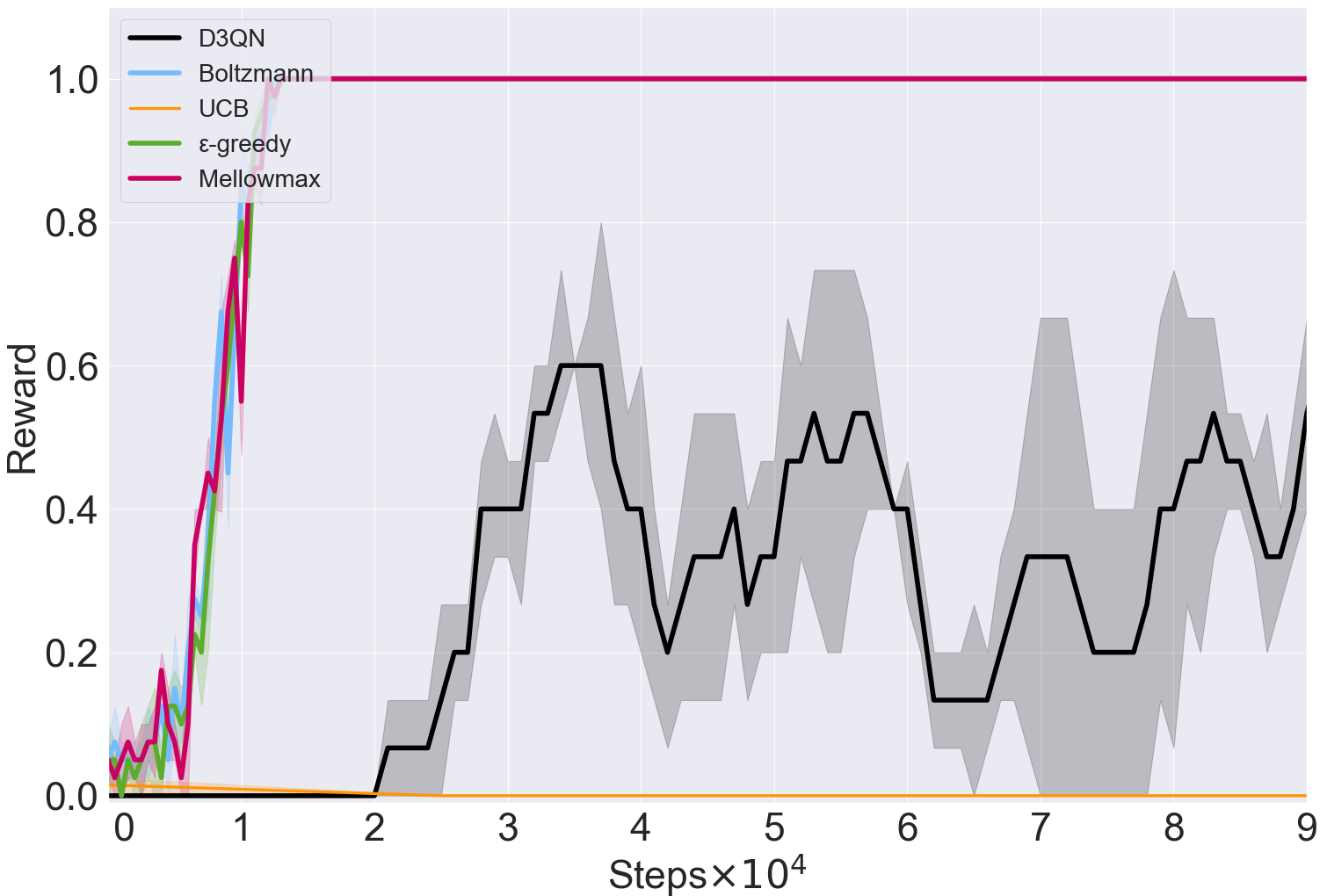}
    \caption{MFEC-based algorithms.}
  \end{subfigure}
  \begin{subfigure}[b]{0.48\textwidth}
    \includegraphics[width=\textwidth]{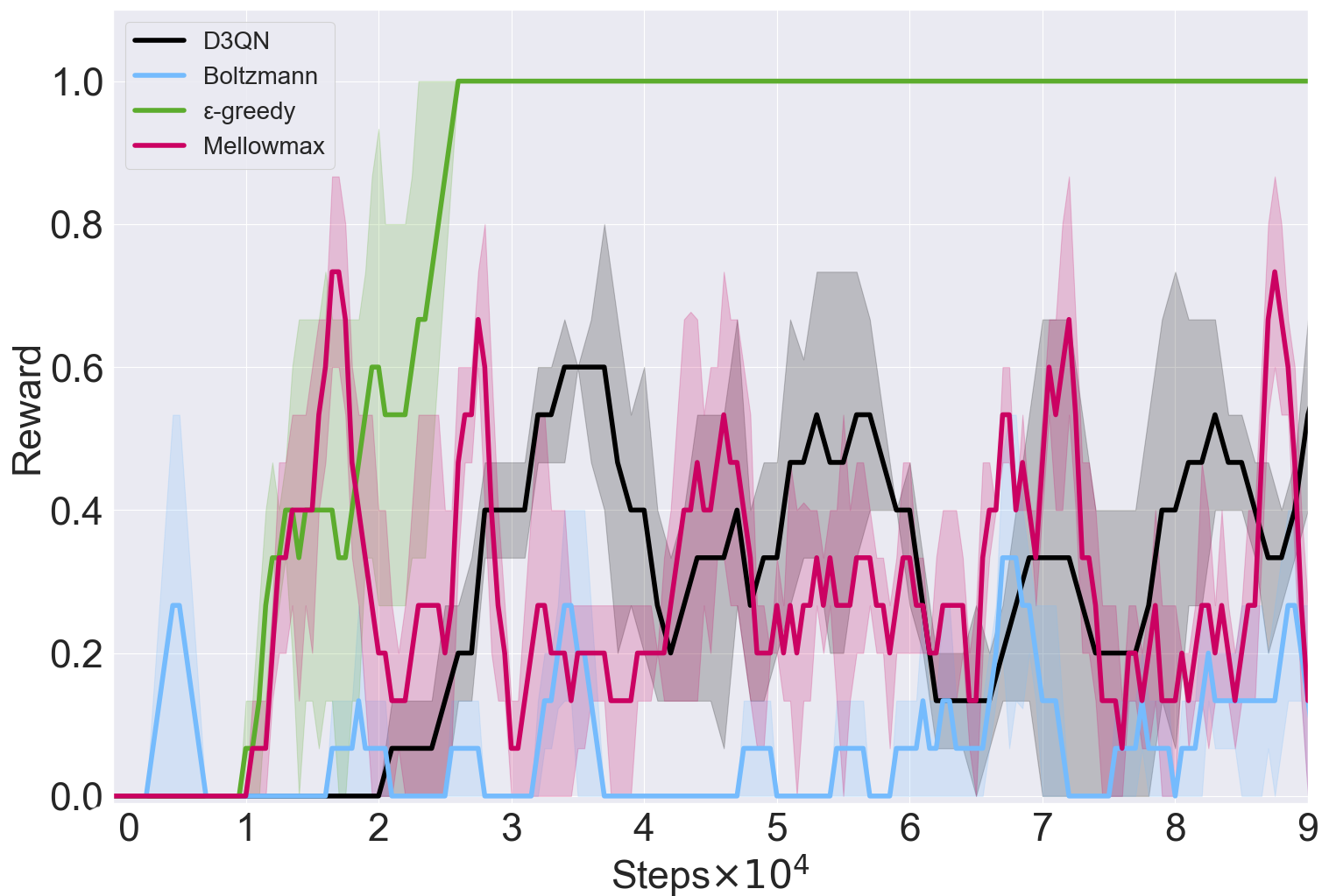}
    \caption{NEC-based algorithms.}
  \end{subfigure}
  \caption{Learning curves on OpenRoom.}
\end{figure}

\space
\begin{figure}[ht]
  \begin{subfigure}[b]{0.48\textwidth}
    \includegraphics[width=\textwidth]{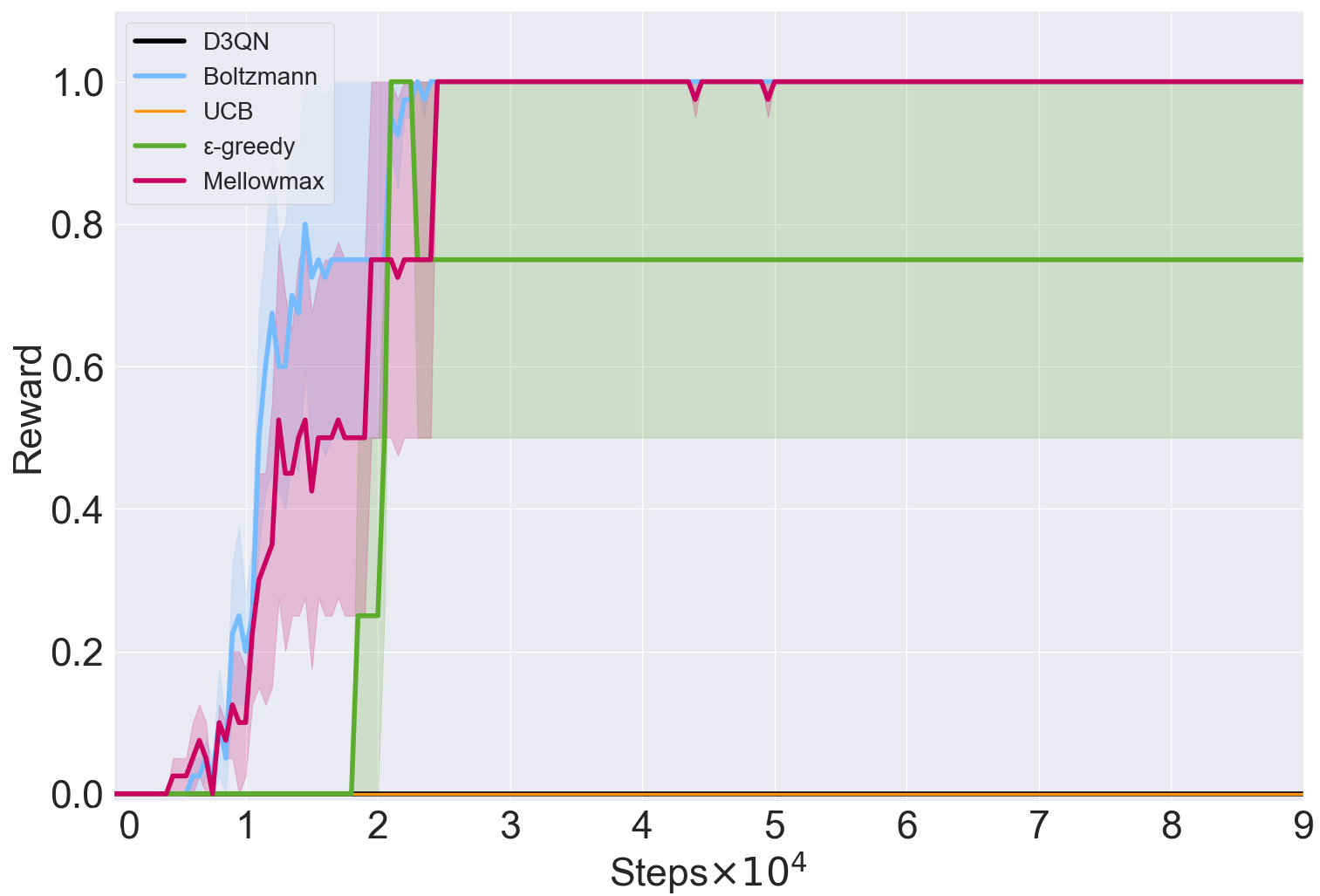}
    \caption{MFEC-based algorithms.}
  \end{subfigure}
  \begin{subfigure}[b]{0.48\textwidth}
    \includegraphics[width=\textwidth]{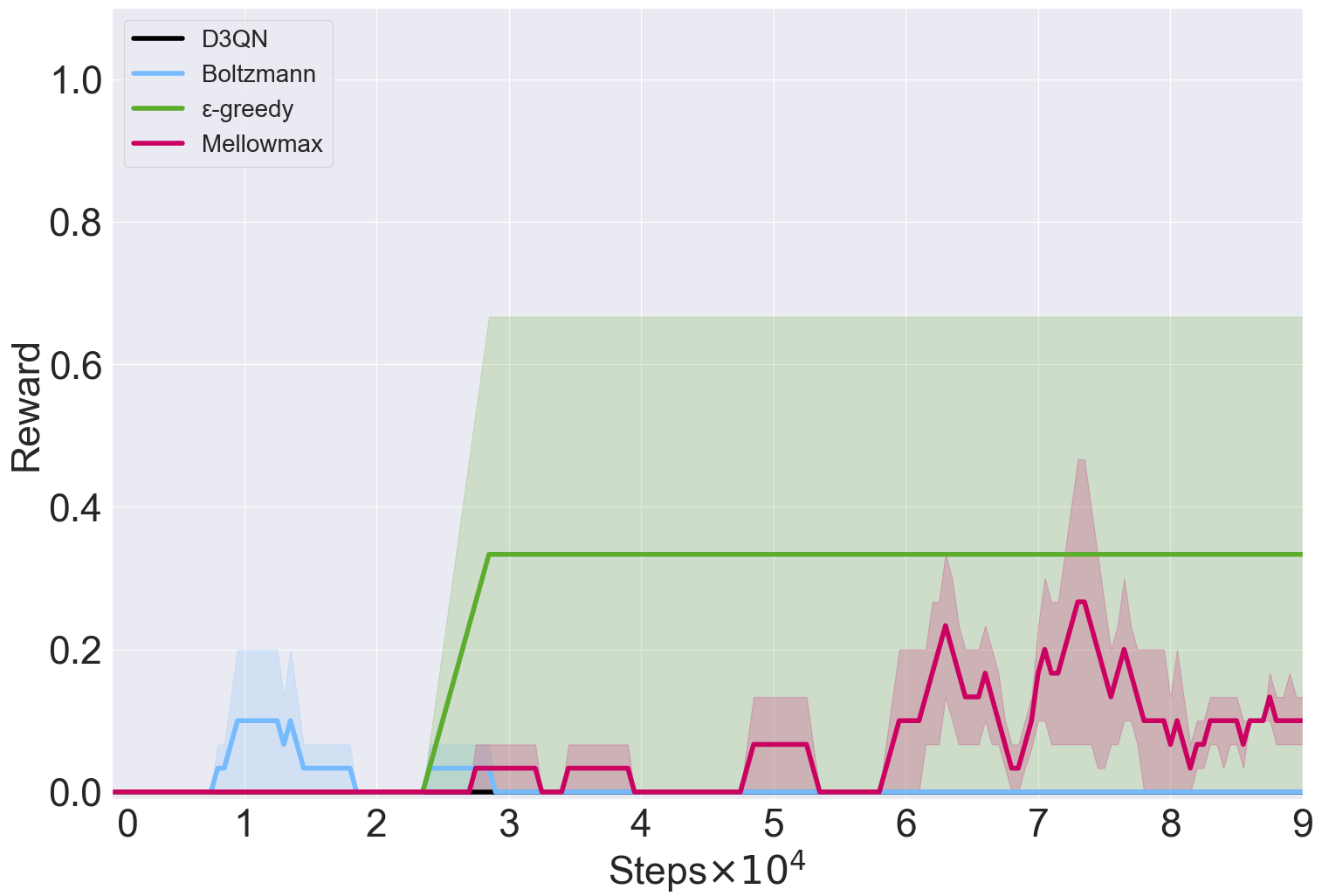}
    \caption{NEC-based algorithms.}
  \end{subfigure}
  \caption{Learning curves on FourRoom.}
\end{figure}

\newpage

\section{Atari Results}
\label{sec:Atari results}

\begin{figure}[ht]
  \begin{subfigure}[b]{0.45\textwidth}
    \includegraphics[width=1\textwidth]{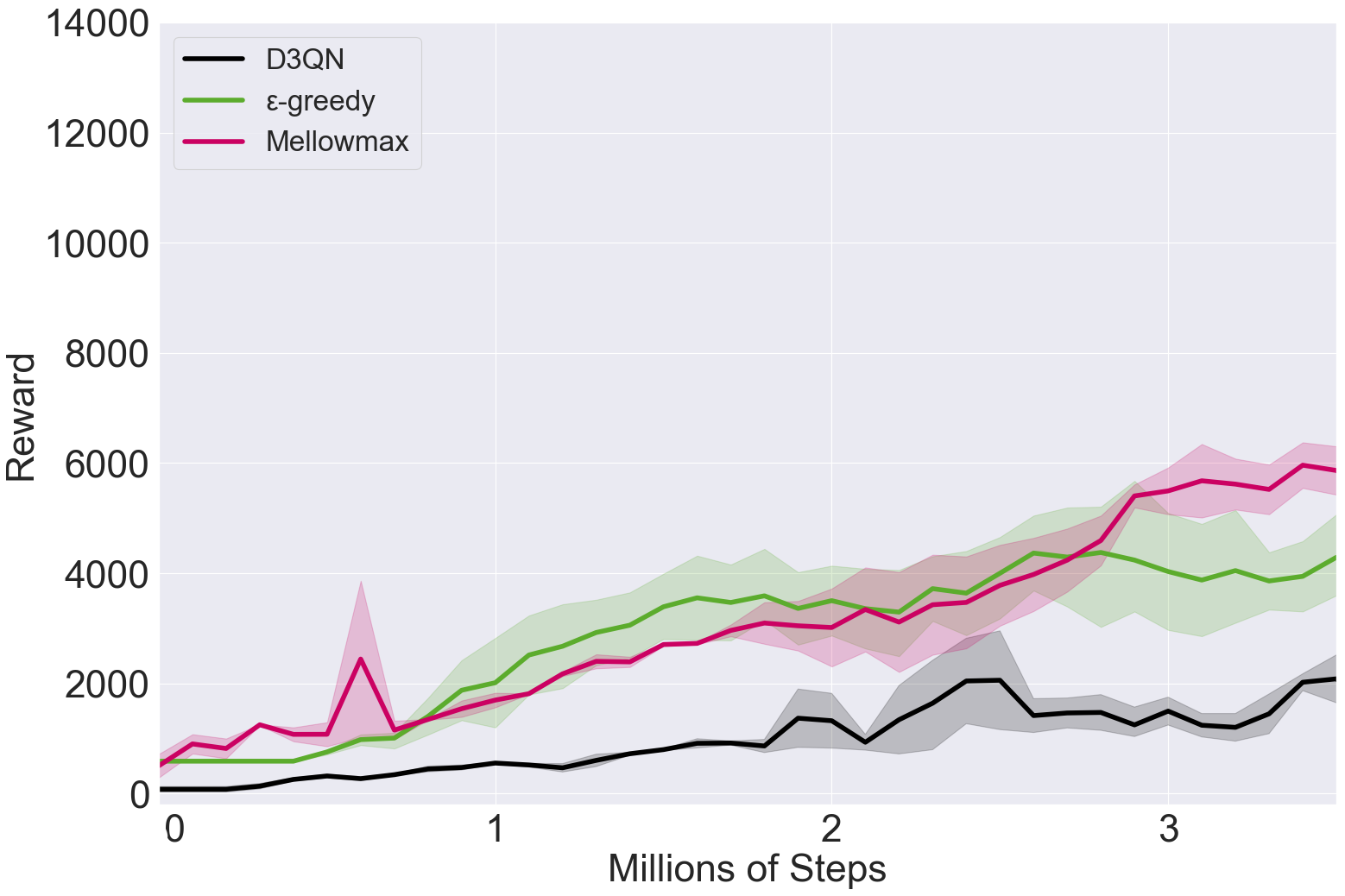}
    \caption{Q*Bert}
  \end{subfigure}\hfill%
  \begin{subfigure}[b]{0.45\textwidth}
    \includegraphics[width=1\textwidth]{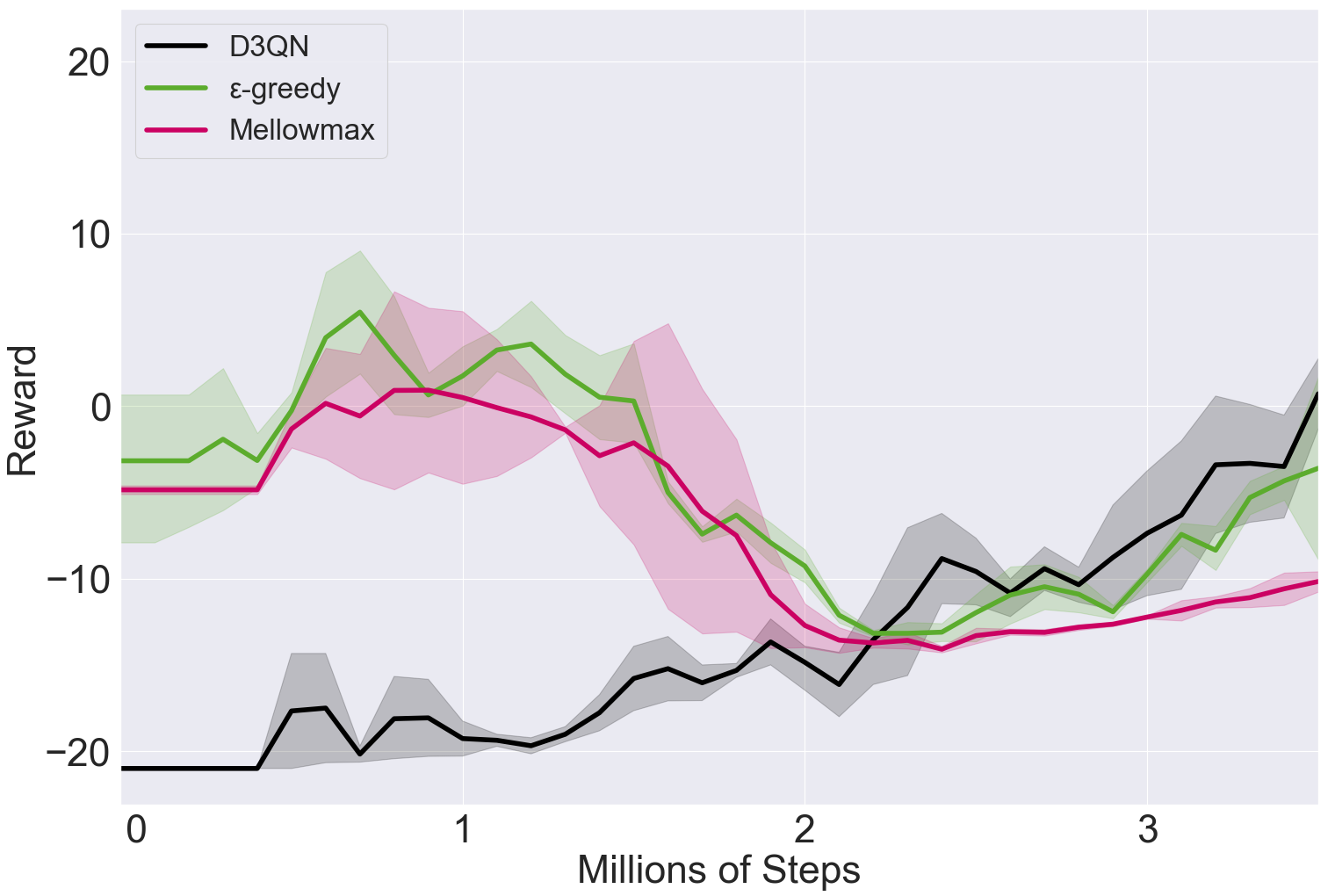}
    \caption{Pong}
  \end{subfigure}\hfill%
  \begin{subfigure}[b]{0.45\textwidth}
    \includegraphics[width=1\textwidth]{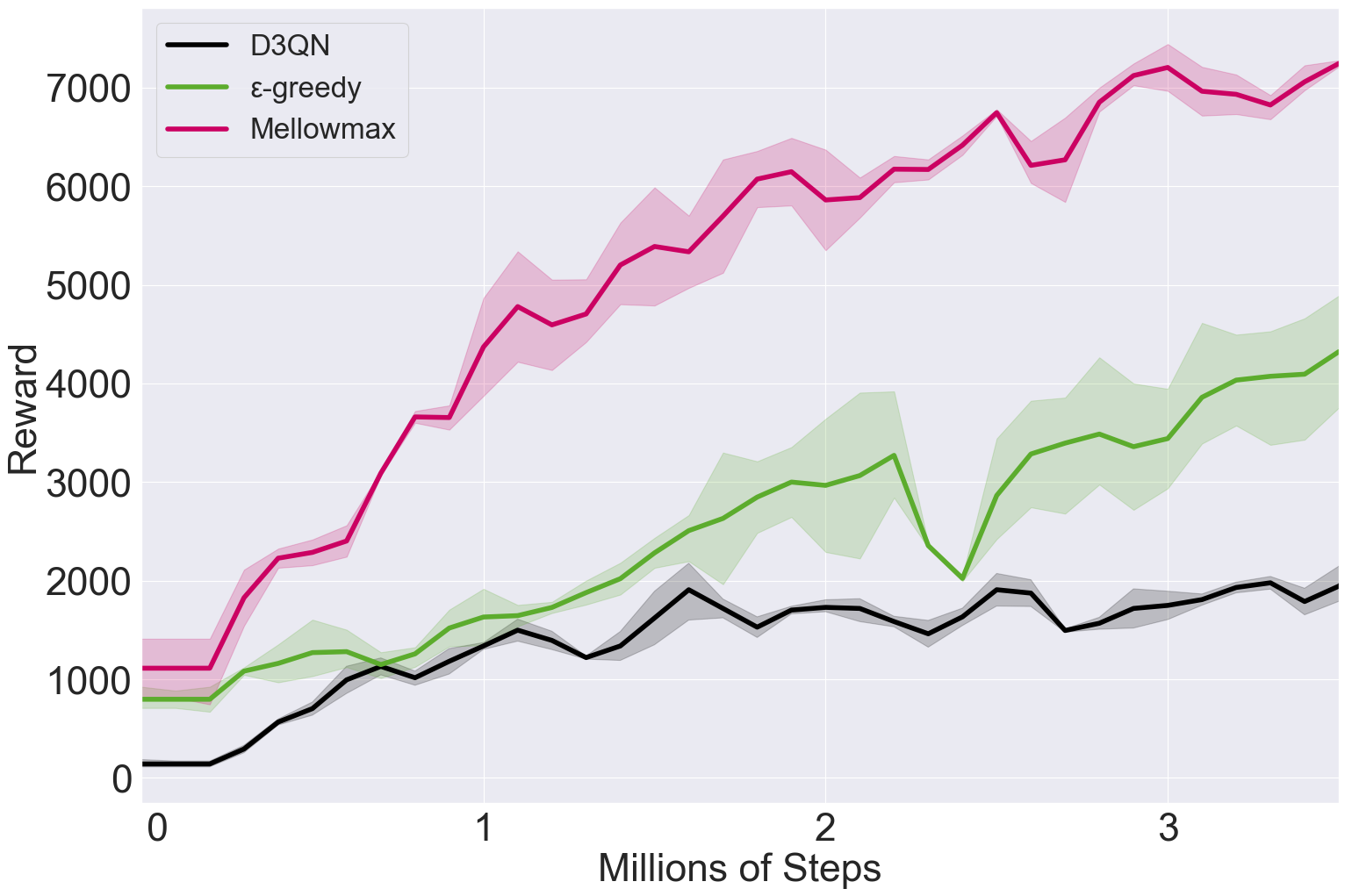}
    \caption{Ms. Pac-Man}
  \end{subfigure}\hfill%
  \begin{subfigure}[b]{0.45\textwidth}
    \includegraphics[width=1\textwidth]{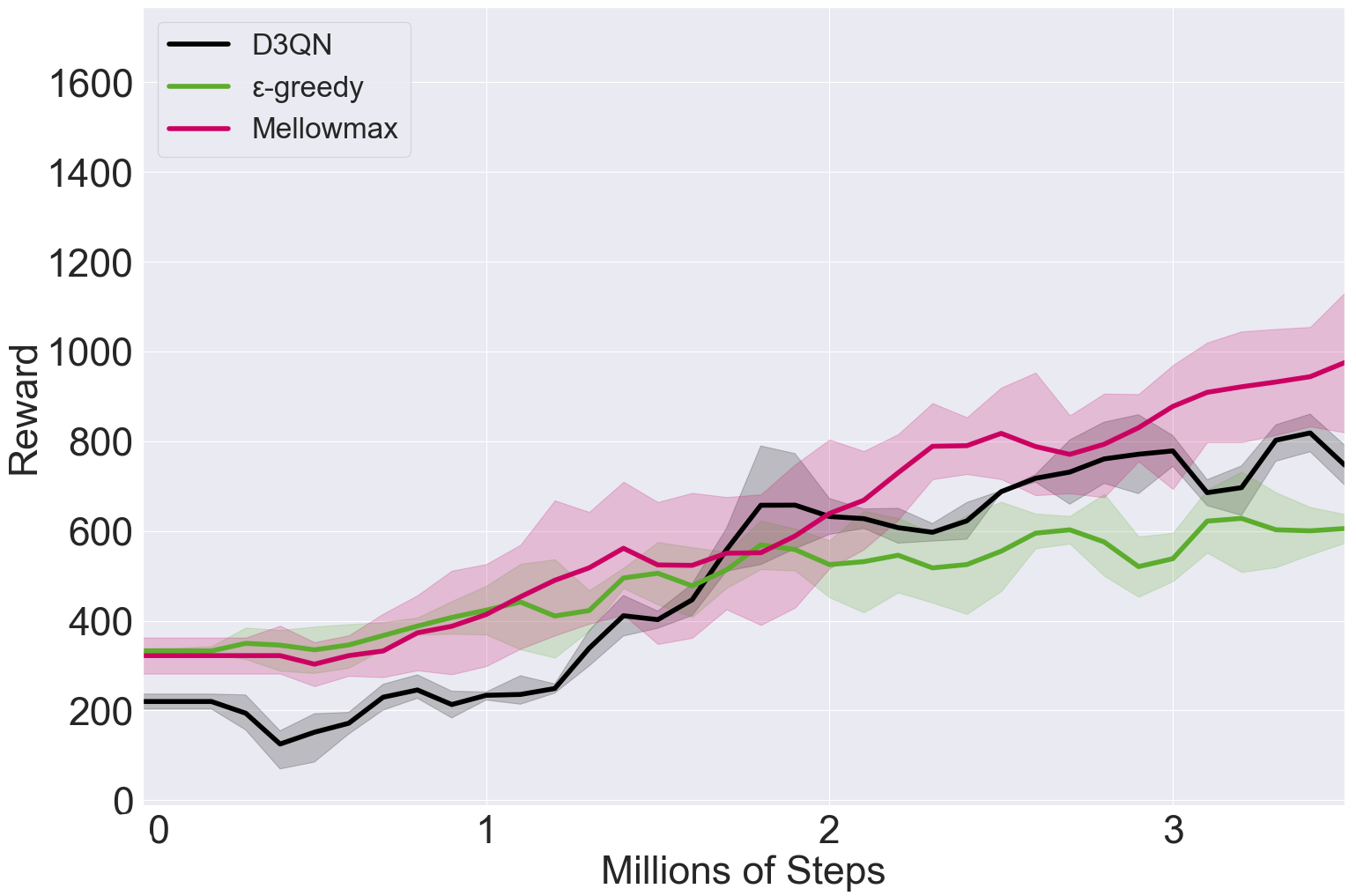}
    \caption{Space Invaders}
  \end{subfigure}\hfill%
  \caption{Learning curves for NEC-based algorithms.}
\end{figure}

\begin{figure*}[ht!]
    \centering
    \includegraphics[width=0.45\textwidth]{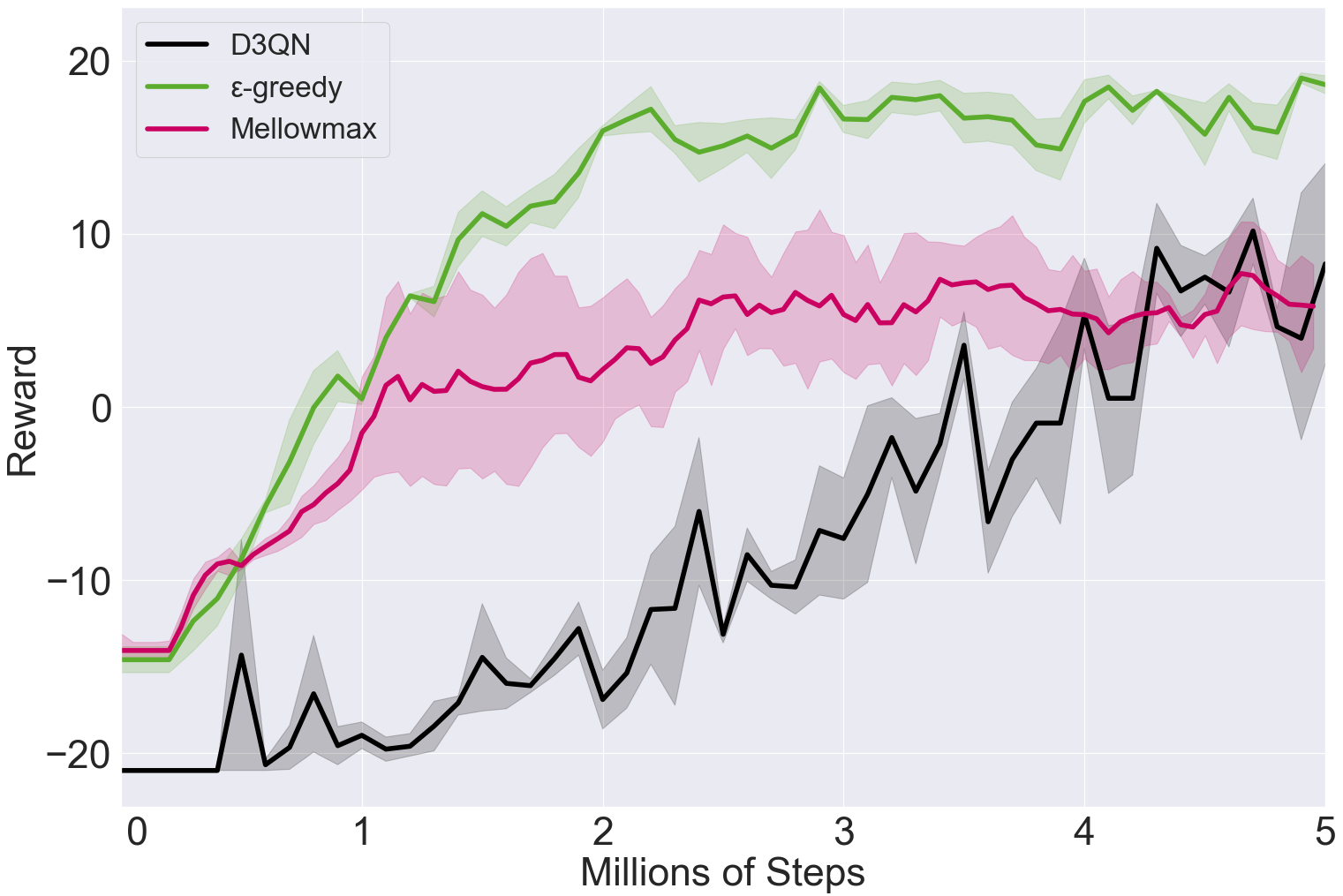}
    \caption{Learning curves for MFEC-based algorithms on Pong.}
\end{figure*}

\clearpage

\section{Hyperparameters}
\label{sec:hyperparameters}

\begin{table}[H]
\centering
\caption{Hyperparameters used for the Gridworld  and the Classic Control Domains.}\smallskip
\begin{tabular}{ c  c  c c  }
\hline
  \textbf{Parameters name} & \multicolumn{1}{c}{\textbf{D3QN}} & \multicolumn{1}{c}{\textbf{MFEC}} & \multicolumn{1}{c}{\textbf{NEC}}\\
 \hline
 \hline
 
 $\epsilon$ initial &$1$&$1$&$1$\\
 
 $\epsilon$ final&$5\times10^{-3}$&$5\times10^{-3}$&$5\times10^{-3}$\\
 
$\epsilon$ anneal start (steps) &$1$&$5\times10^3$&$5\times10^3$\\
 
$\epsilon$ anneal end (steps)&$5\times10^{4}$&$25\times10^3$&$25\times10^3$\\
 
 Discount factor $\lambda$&$0.99$&$0.99$&$0.99$\\
 
 Reward clip &Yes&No&No\\

 Number of neighbours $k$&---&$11$&$11$\\ 
 
 Kernel delta $\delta$&---&$10^{-3}$&$10^{-3}$\\
 
 Experience replay size &$10^5$&---&$10^5$\\

RMSprop learning rate &$25\times10^{-5}$&---&$7.92\times10^{-6}$\\

RMSprop momentum &$0.95$&---&$0.95$\\

RMSprop $\epsilon$ &$10^{-2}$&---&$10^{-2}$\\

Training start (steps)&$5\times10^3$&---&$10^3$\\

Batch size&$32$&---&$32$\\

Target network update (steps) &$7.5\times10^3$&---&---\\

Observation projection &---& None& ---\\
 
Projection key size & ---& None& ---\\

Memory learning rate $\alpha$ &---&---&$0.1$\\

$n$-step return&---&---&$100$\\

Key size &---&---&$64$\\

 \hline
\end{tabular}

\label{control_gridworld}
\end{table}

\begin{table}[H]
\centering
\small
\caption{Hyperparameters used for the Atari games.}\smallskip
\begin{tabular}{ c  c  c c  }
\hline
  \textbf{Parameters name} & \multicolumn{1}{c}{\textbf{D3QN}} & \multicolumn{1}{c}{\textbf{MFEC}} & \multicolumn{1}{c}{\textbf{NEC}}\\
 \hline
 \hline
 
 $\epsilon$ initial &$1$&$1$&$1$\\
 
 $\epsilon$ final&$10^{-2}$&$5\times10^{-3}$&$10^{-3}$\\
 
$\epsilon$ anneal start (steps) &$1$&$5\times10^3$&$5\times10^3$\\
 
$\epsilon$ anneal end (steps)&$10^{6}$&$25\times10^3$&$25\times10^3$\\
 
 Discount factor $\lambda$&$0.99$&$0.99$&$0.99$\\
 
 Reward clip &Yes&No&No\\

 Number of neighbours $k$&---&$11$&$11$\\ 
 
 Kernel delta $\delta$&---&$10^{-3}$&$10^{-3}$\\
 
 Experience replay size &$10^6$&---&$10^5$\\

RMSprop learning rate &$25\times10^{-5}$&---&$7.92\times10^{-6}$\\

RMSprop momentum &$0.95$&---&$0.95$\\

RMSprop $\epsilon$ &$10^{-2}$&---&$10^{-2}$\\

Training start (steps)&$12.5\times10^3$&---&$5\times10^4$\\

Batch size&$32$&---&$32$\\

Target network update (steps) &$10^3$&---&---\\

Observation projection &---& Gaussian& ---\\
 
Projection key size & ---& 128& ---\\

Memory learning rate $\alpha$ &---&---&$0.1$\\

$n$-step return&---&---&$100$\\

Key size &---&---&$128$\\

 \hline
\end{tabular}

\end{table}

\end{document}